\def\year{2021}\relax
\documentclass[letterpaper]{article} 
\usepackage{aaai21}  
\usepackage{times}  
\usepackage{helvet} 
\usepackage{courier}  
\usepackage[hyphens]{url}  
\usepackage{graphicx} 
\usepackage{natbib}  
\usepackage{caption} 
\usepackage{booktabs}
\usepackage{amssymb}
\usepackage{hyperref}

\title{Boundary-Aware Geometric Encoding for Semantic Segmentation of Point Clouds}
\author{
    Jingyu Gong,\textsuperscript{\rm 1$*$}
    Jiachen Xu, \textsuperscript{\rm 1}\thanks{Equal Contribution.}
    Xin Tan, \textsuperscript{\rm 1,3}
    Jie Zhou, \textsuperscript{\rm 3}
    Yanyun Qu, \textsuperscript{\rm 4}
    Yuan Xie, \textsuperscript{\rm 2$\dagger$}
    Lizhuang Ma \textsuperscript{\rm 1,2}\thanks{Corresponding Author.} \\
}
\affiliations{
    \textsuperscript{\rm 1} Department of Computer Science and Engineering, Shanghai Jiao Tong University, Shanghai, China \\
    \textsuperscript{\rm 2} School of Computer Science and Technology, East China Normal University, Shanghai, China \\
    \textsuperscript{\rm 3} City University of Hong Kong, HKSAR, China \\
    \textsuperscript{\rm 4}  School of Informatics, Xiamen University, Fujian, China \\
    \{gongjingyu, xujiachen, tanxin2017\}@sjtu.edu.cn, jzhou67-c@my.cityu.edu.hk \\
    yyqu@xmu.edu.cn, yxie@cs.ecnu.edu.cn, ma-lz@cs.sjtu.edu.cn
}

\begin{document}

\maketitle

\begin{abstract}
Boundary information plays a significant role in 2D image segmentation, while usually being ignored in 3D point cloud segmentation where ambiguous features might be generated in feature extraction, leading to misclassification in the transition area between two objects. In this paper, firstly, we propose a Boundary Prediction Module (BPM) to predict boundary points. Based on the predicted boundary, a boundary-aware Geometric Encoding Module (GEM) is designed to encode geometric information and aggregate features with discrimination in a neighborhood, so that the local features belonging to different categories will not be polluted by each other. To provide extra geometric information for boundary-aware GEM, we also propose a light-weight Geometric Convolution Operation (GCO), making the extracted features more distinguishing. 
Built upon the boundary-aware GEM, we build our network and test it on benchmarks like ScanNet v2, S3DIS. Results show our methods can significantly improve the baseline and achieve state-of-the-art performance. Code is available at \href{https://github.com/JchenXu/BoundaryAwareGEM}{https://github.com/JchenXu/BoundaryAwareGEM}.
\end{abstract}

\section{Introduction}

\begin{figure*}[ht]
     \centering
     \includegraphics[width=\linewidth]{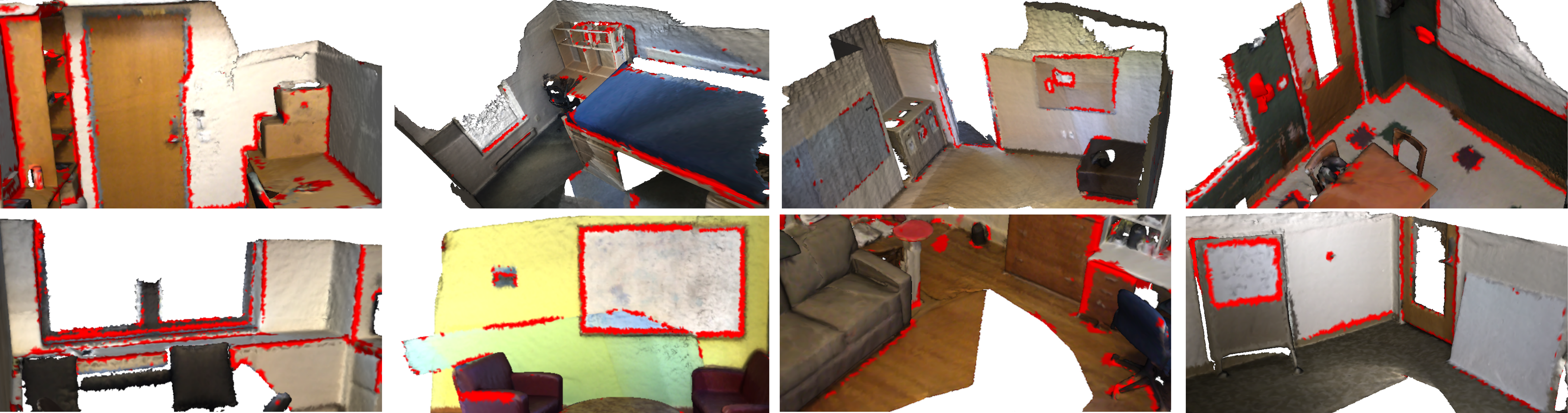}
     \caption{Visualization of boundary predicted by our proposed Boundary Prediction Module (BPM). Points in red represent the predicted boundary points. Each visualized scene is selected from ScanNet v2.}
     \label{fig:boundary-visual}
\end{figure*}

Semantic segmentation of point clouds has become an increasingly attended task. Because of the success of 2D image recognition~\cite{long2015fully,chen2017deeplab}, many works tried to extend 2D convolution network to 3D space directly~\cite{maturana2015voxnet,zhou2018voxelnet}. However, this kind of methods is limited by drastic increment of computational complexity. On the other side, PointNet~\cite{qi2017pointnet} utilized shared Multi-Layer Perceptrons to directly process point clouds and aggregates information through max-pooling, but it failed to exploit the relationship among points in a local region. Due to the unbalanced distribution of points and irregularity of representation, semantic segmentation of point clouds is still a challenging task.

The boundary plays an important role in the semantic segmentation of point clouds, because lots of misclassifications happen nearby boundary points. In the point cloud, the boundary refers to the transition area between two or more objects belonging to different categories. For example, the junction of the sofa and the ground can be considered as the boundary. Many works~\cite{wang2018deep,xu2018spidercnn,wu2019pointconv} tackled the segmentation problem in point clouds without explicitly learning or using the boundary information, hence they extracted features from points with no differentiation between boundary and non-boundary points. It is noteworthy that extracted features on the boundary are usually ambiguous, because they mix features of points belonging to different categories on different sides of the boundary. As the network goes deeper, if other points incorporate features of the boundary points, these ambiguous features on the boundary will inevitably propagate to more other points hierarchically. So, the information of different objects will spread across the boundary, leading to a bad contour for final semantic segmentation.

To tackle this problem, we propose a Boundary Prediction Module (BPM) to predict boundary points in point clouds. In this module, we give a soft prediction for boundary and this module is skillfully supervised by the ground truth of boundary generated on the fly. It is noteworthy that, compared with semantic segmentation, boundary prediction is easier and likely to obtain better results. So, we introduce the light-weight BPM to predict the boundary. Then, we use the prediction as auxiliary information to boost the performance of segmentation. The BPM and segmentation network are trained jointly in end-to-end manner. Fig. \ref{fig:boundary-visual} illustrates the predicted boundary in several scenes. Most of them are accurately located between different categories, which also visually reflects the effectiveness of our BPM.

Based upon the BPM, we design a boundary-aware Geometric Encoding Module (GEM) to utilize the predicted boundary in feature extraction. When aggregating local features, we only allow information sharing within each object area by preventing the propagation of features across boundary. Because local features can provide more detail information, mixing local features of different categories will definitely destroy this detail information. Then, in the following layers of encoder where representative points are sampled and global features are encoded, information belonging to different categories can be transferred through boundary to obtain the global scene information.
In this way, the predicted boundary would act as a barrier to prevent the information mixture from different categories in local feature extraction and be ignored in global feature extraction.

To effectively exploit geometric information, we design a light-weight Geometric Convolution Operation (GCO) which complements geometric features for the boundary-aware GEM. In the GCO, we focus on the angular distribution of neighbors rather than the spatial distribution used in KCNet~\cite{shen2018mining} and KPConv~\cite{thomas2019kpconv}, which is sensitive to density of points and lack of generalization. In specific, we use a simple vector set as the trainable kernel to learn the geometric pattern. In a neighborhood with $m$ points, its geometric pattern can be represented by $m$ 3-D directional vectors. Therefore, our proposed trainable geometric kernel has the same form. Then, the geometric convolution is a sum over multiplication of vectors in the kernel and directional vectors in the neighborhood. Like the 2D convolution, the response of GCO will be large if the local geometric pattern is similar to the learnt kernel. 

Overall, the major contributions can be summarized as follows: (1) We propose a boundary-aware Geometric Encoding Module (GEM) to accurately encode geometric information and prevent the propagation of information across the boundary in local feature extraction. To our best knowledge, we are the first one to take boundary information into the 3D feature aggregation process in an explicit way. (2) The Boundary Prediction Module (BPM), which is supervised with dynamically generated ground truth, is derived to predict the boundary and provide boundary information for the boundary-aware GEM. (3) A Geometric Convolution Operation (GCO) with a learnable vector kernel set is also designed to explore local geometry for each point in a light-weight manner. Experiments on benchmark datasets show that the cutting-edge backbone with the proposed boundary-aware GEM can achieve the state-of-the-art performance.

\section{Related Work}
\label{sec:related}

\subsubsection{Point cloud semantic segmentation.}
\label{subsec:seg_related}
Intuitively, voxel-based methods~\cite{maturana2015voxnet,zhou2018voxelnet} voxelized point clouds and applied 3D grid convolution. Furthermore, SubSparseConv~\cite{graham20183d} proposed a convolution for sparse point clouds. However, voxelization inevitably destroys geometric information. PointNet~\cite{qi2017pointnet} directly extracted features from the point cloud through shared Multi-Layer Perceptrons. Then, PointNet++~\cite{qi2017pointnet++} introduced a hierarchical network to aggregate information from a local region and extract features from different scales. But they merely used max-pooling to aggregate information, not typically considering the spatial convolution.

To simulate the spatial convolution operation used in image processing, PCCN~\cite{wang2018deep} and SpiderCNN~\cite{xu2018spidercnn} utilized MLPs and 3-order function, respectively, to approximate to 3D continuous weight functions w.r.t the position. To deal with the unbalanced distribution of point clouds, PointConv~\cite{wu2019pointconv} estimated the density of every point and re-balanced the contribution of each point during convolution according to the point density. PointASNL~\cite{yan2020pointasnl} also re-weighted the neighbors to adjust the location of sampled centering point. Additionally, HEPIN~\cite{jiang2019hierarchical} introduced an edge branch to exploit the relationship between neighbors and collaborated it with the main branch for fine-grained context information. SPH3D-GCN~\cite{lei2020spherical} proposed a spherical convolutional kernel splitting the neighborhood into multiple volumetric bins.

All these work tried to simulate the 3D convolution operation aggregating information from a local region without differentiation among points. Compared with these methods, our method is boundary-aware to treat points differently for feature aggregation in a local region, so as to alleviate the propagation of indistinguishable features.

\subsubsection{Boundary in semantic segmentation.}
\label{subsec:bound_related}
The convolution operation in a local region will aggregate information from neighbors no matter which category they belong to, making the extracted features ambiguous on the boundary, as the neighborhood may include objects belonging to different categories on different sides of boundary. GAC~\cite{wang2019graph} determined the weight of every point's feature according to the similarity, thus alleviating the ambiguity of features introduced by aggregating features of points with different labels. But, no boundary information is explicitly involved, leading to a sub-optimal result. However, the boundary information is quite useful for high-level vision task like semantic segmentation~\cite{bertasius2015high}. To enhance segmentation coherence, BNF~\cite{bertasius2016semantic} used a combination of feature maps to predict the boundary and defined a boundary pairwise potential for energy minimization. BSANet~\cite{zhao2019multi} detected the boundary in images and emphasized the features near the boundary at the early stages. These methods proved the importance of boundary for the task of semantic segmentation. 

In this work, we also propose the Boundary Prediction Module (BPM) to predict the boundary for point clouds and adjust the feature propagation of boundary points. Compared with BSANet~\cite{zhao2019multi}, we (1) suppress the propagation of point features on the boundary and (2) predict the boundary for the input point clouds and sample boundary points for other scales.

\subsubsection{Geometric features in point clouds.}
\label{subsec:geo_related}
Compared with 2D image, point clouds provide more geometric information. ShapeNet~\cite{chang2015shapenet} provided the normal vector along with $xyz$ for every point. PPF-FoldNet~\cite{deng2018ppf} designed geometric features based on the angles between relative positions and normal vectors, but the features themselves were not learnable. KCNet~\cite{shen2018mining} proposed to use a learnable point-set kernel to represent the geometric pattern. Specifically, they utilized Gaussian kernel with the distances between kernel points and anchor points as the input to obtain the similarity between the point-set kernel and neighbors distribution.

Similary to KCNet, our kernel is also a set of vectors. However, to extract geometric features, our proposed Geometry Convolution Operation (GCO) focuses on the direction rather than the position in the kernel, thus less sensitive to sampling density of points. Besides, our GCO is light-weight and extracts features hierarchically to learn effective geometric patterns, rather than using a heavy-weight module to extract the geometric features just in one layer.

\section{Methods}
\label{sec:method}
First, we will introduce the overall architecture. Second, we will show how we detect the boundary and describe the proposed boundary-aware geometric encoding in detail. Finally, the geometric convolution, which is simply designed but extracts the geometric information efficiently, will be introduced in detail.

\subsection{Network Overview}
\label{subsec:overview}
\begin{figure*}[t]
    \centering
    \includegraphics[width=0.9\linewidth]{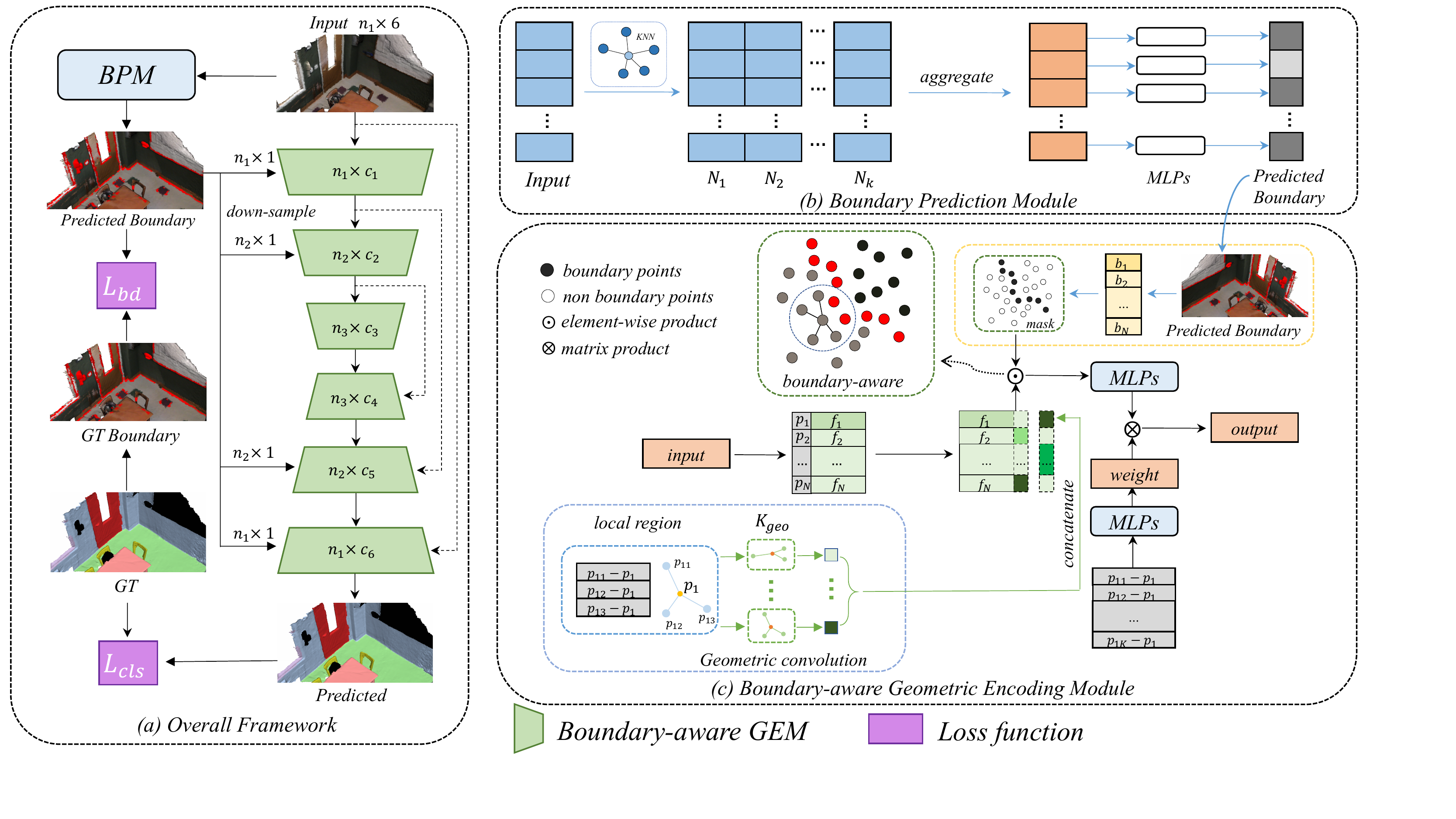}
    \caption{Overall architecture of our network consisting of the Boundary Prediction Module and the boundary-aware Geometric Encoding Module for each layer. (a) illustrates the overall architecture. (b) depicts the architecture of Boundary Prediction Module. (c) describes the boundary-aware Geometric Encoding Module in details.}
    \label{fig:framework}
\end{figure*}

In this paper, we fully consider the geometric characteristics of scenes. Overall, as shown in Fig. \ref{fig:framework} (a), we propose an encoder-decoder network composed of a Boundary Prediction Module (BPM) and boundary-aware Geometric Encoding Module (GEM). The BPM is a small and concise neural network to predict the boundary points, so as to provide the boundary cues for boundary-aware GEM to adjust the feature propagation in local regions.

Meanwhile, the boundary-aware GEM also encodes the geometric information of the local region with the help of new derived Geometric Convolution Operation (GCO), that will be described later. It is noteworthy that boundary is only involved when the number of points is large (i.e., the early stage of encoder and the later stage of decoder). In other layers, all points are treated as the non-boundary points and we only focus on the geometric context.

\subsection{Boundary-Aware Geometric Encoding}
\label{subsec:boundary}

To implement the boundary-aware GEM, we first introduce a Boundary Prediction Module to predict the boundary points given the point cloud. This module is regularized by the target boundary generated on the fly based on semantic labels. Later, the predicted boundary information is used to impede the propagation of information across the boundary for local feature extraction. By contract, global and abstract features can cross over the boundary to have a better recognition of the global scene.

\subsubsection{Boundary Prediction Module.}
First, we automatically annotate each point in training samples as its indicator of the boundary $g$, which is defined in accordance with the label of every point as below. In the target boundary, $g_i$ is $0$ if the $i^{th}$ point is on the boundary, otherwise equal to $1$. For every point $p$, whether it is located on the boundary is determined by its local neighborhood. That is, given fixed number of neighboring points for $p$, if there are more than a predefined ratio (detailed description is in experiments) of points that do not belong to the same category as $p$, then $p$ is assumed to be the point on the boundary, otherwise it is not.

The boundary prediction task is a slightly different from semantic segmentation, as boundary prediction should be aware of the difference of semantic information in a local region. To this end, as shown in Fig. \ref{fig:framework}. (b), we collect features of $k$ nearest neighbors in the local region for each point and take the variance of collected features as the input of the following part of BPM. Then, like PointNet~\cite{qi2017pointnet}, we utilize several shared MLPs to predict the boundary annotation $\hat{g}$ for the whole input point cloud. Compared with a carefully designed network, our BPM is compact and easy to train. Specifically, its training loss is following:

\begin{equation}
    \mathcal{L}_{BPM}=-\sum_{i=1}^n(w_1\cdot  g_i\log\hat{g_i}+w_2\cdot (1-g_i)\log(1-\hat{g_i})),
\end{equation}
where $w_1$ and $w_2$ are used to balance the huge difference between the numbers of two categories. We also utilize cross-entropy loss to regularize the final semantic segmentation output, and the total loss is a simple addition of boundary prediction loss and semantic segmentation loss.

\subsubsection{Feature Aggregation with Boundary.}
As mentioned above, in the proposed boundary-aware GEM (Fig. \ref{fig:framework} (c)), we attempt to block the propagation of local features from points on the boundary in the early stage of encoding process. Therefore, according to the predicted boundary, we utilize the boundary information as a mask/filter to assign different weights to different points during feature aggregation. Before that, we also utilize the GCO (detailed description will be given later) to provide extra geometric features. The main difference between boundary-aware GEM decoder and encoder is that, we do not use the GCO in decoder as the output features of corresponding encoder which have already contained the geometry information will be concatenated to the input of the decoder.

Given the predicted boundary points (the red points in Fig. \ref{fig:framework}. (c)), during feature aggregation for a grey point, it will collect features in a neighborhood but ignore those points on the boundary. Therefore, the local feature aggregation for point $p_i$ can be expressed as follows:
\begin{equation}\label{local_aggregation}
    f_{en\_\,l}=\sigma(\mathcal{A}(\{\phi(r_{ij})\cdot\mathcal{M}(\hat{g_j}\cdot f_{p_j})\})),\ \forall p_j\in\mathcal{N}(p_i) ,
\end{equation}
where $f_{p_j}$ represents the feature of neighboring $p_j$ containing both original features and geometric features, and $\hat{g_j}$ works as a mask to assign weight to $f_{p_j}$. In this formula, $\mathcal{N}(p_i)$ is the neighborhood of $p_i$, and $\mathcal{M}$ means shared MLPs to combine the original features and extracted geometric features at this scale. Referring to Fig. \ref{fig:framework} (c), we can know $\phi$ learns weight from the relative position $r_{ij}$ for neighbor $p_j$ through another few MLPs. Additionally, $\mathcal{A}$ is the aggregation function that is done through matrix product and $\sigma$ represents the activation function. It is noteworthy that $\hat{g_j}$ is 0 if $p_j$ is on the boundary and this boundary point would not contribute to the aggregated feature.

In this way, we prevent the features of points on the boundary to be fused into the extracted local features, thus information is less likely to cross over the boundary to pollute features belonging to other categories (shown in Fig. \ref{fig:framework}. (c)). We only need to predict the boundary of point clouds for the input layer, while in the later encoding stages, points and predicted boundary labels are down-sampled at the same time. Unlike local features in the first few layers, the global features can propagate among different objects through boundary points. Therefore, in the latter stage, we extract global features as follows:

\begin{equation}
    f_{en\_\,h}=\sigma(\mathcal{A}(\{\phi(r_{ij})\cdot \mathcal{M}(f_{p_j})\})),\ \forall p_j\in\mathcal{N}(p_i).
\end{equation}

In the decoding stage, the feature extraction procedure is symmetrical. Specifically, when the number of points remains small, global features propagate without impeding to better recognize the global context. While in the later stage of the decoder, we prevent the propagation of features across the boundary again to obtain distinguishing local features.

\subsection{Geometric Convolution}
\label{subsec:geometry}
To provide extra geometric information for boundary-aware GEM, we propose a light-weight Geometric Convolution Operation (GCO) with a learnable kernel to extract geometric information at different scales, see the bounding box on the left lower corner in Fig. \ref{fig:framework} (c). 

\subsubsection{Geometric Kernel.}

\begin{figure}[t]
    \centering
    \includegraphics[width=\linewidth]{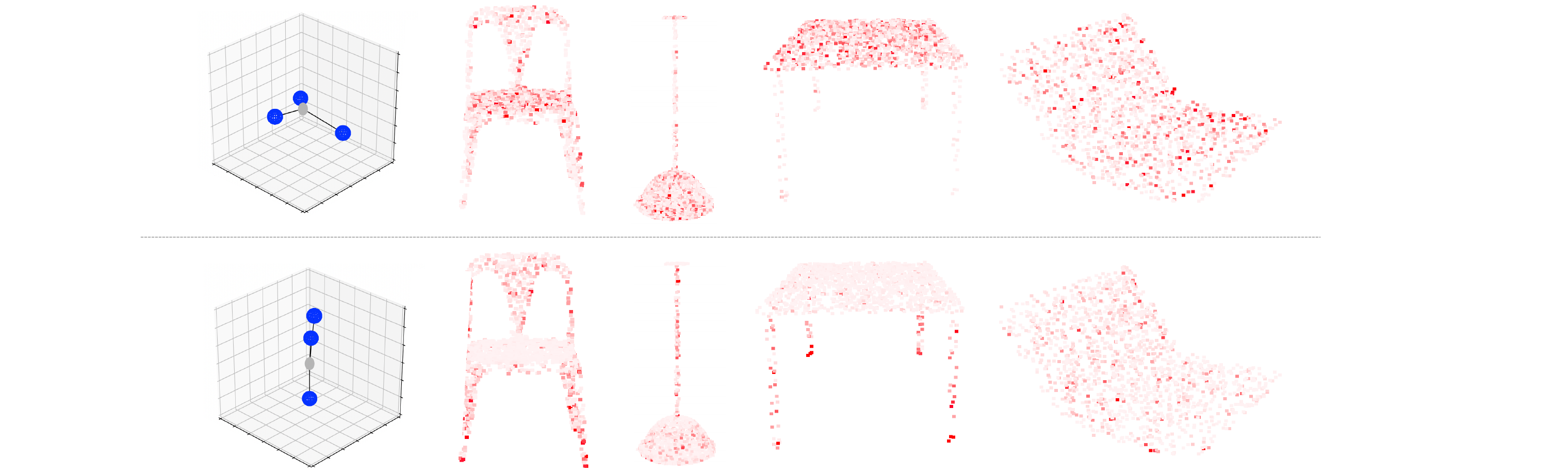}
    \caption{Heat maps for the response of different learnt kernels. The left-most column shows the kernels and the right columns shows the response of some examples (the redder, the larger). These two kernels separately learn the pattern of a horizontal plane and a vertical line.}
    \label{fig:kernel}
\end{figure}

In our method, we propose a geometric kernel $K_{geo}$ with three directional vectors $\{v_1, v_2, v_3 | v_i \in \mathbb{R}^3\}$. Each vector represents a direction in the 3D space, thus the kernel itself can describe a distribution of points over directions, so as to tell where the points is located (e.g., on a plane or curved surface). Unlike \cite{shen2018mining,thomas2019kpconv} which employ a large amount of kernel points, only three 3-D directional vectors are adopted in our method. Even though the proposed operation has a much simpler structure, the performance is comparable with some sophisticated operators,  that is proved in ablation study. Because, tetrahedron is the simplest polyhedron and these three directional vectors along with the origin can represent a tetrahedron. Furthermore, more complex geometry pattern can be recognized through hierarchical geometric feature extraction. Fig. \ref{fig:kernel} illustrates learnt kernels and heat maps for different objects to show effectiveness.

\subsubsection{Geometric Convolution Operation.}
For a point in the point cloud, the local pattern is represented by the relative positions from this point to its neighbors. Similar to 2D convolution, if the geometric pattern of the neighborhood is very similar to the learnt GCO kernel, the response will be large, thus geometric pattern is recognized.

Our geometric convolution focuses more on the angular distributions of neighbors rather than their relative displacement like KCNet~\cite{shen2018mining}. For every point $p_i$, relative positions of three neighbors, that are used to represent the local pattern, are represented by $\{\vec{d}_{ij} | \ j\in[1,2,3]\}$. Convolving with $K_{geo}$, the output could be expressed as 
\begin{equation}
    O_i = \mathop{\max}_{\mathcal{P}_i}\sigma(b + \mathop{\sum}_{j}\vec{d}_{ij} \cdot \vec{v}_{{\mathcal{P}_i}(j)}),
\end{equation}
where $b$ is the bias and $\sigma$ is the activation function. $\mathcal{P}_i(\cdot):\{1, 2, 3\}\mapsto\{1, 2, 3\}$ represents a mapping function which finds the matching vector in the kernel for $\vec{d}_{ij}$. It is noteworthy that because the point cloud is unordered, it is hard to use a fixed mapping. Additionally, if $K_{geo}$ describes the same pattern as the neighborhood, each pair of $\vec{d}_{ij}$ and the matching $\vec{v}_{{\mathcal{P}_i}(j)}$ would be in the same direction making the dot production maximum. Therefore, in our proposed convolution procedure, we dynamically choose the mapping function that makes the output maximum. Obviously, our geometric convolution is more sensitive to the angle between two vectors $cos\langle \vec{d}_{ij}, \vec{v}_{{\mathcal{P}_i}(j)}\rangle$ rather than the displacement between vectors in the neighborhood and kernel $\vert \vec{d}_{ij} - \vec{v}_{{\mathcal{P}_i}(j)}\vert$, which is more easily be influenced by scales and density of point clouds.

After extracting geometric features, they are concatenated to the original features of points for further boundary-aware geometric encoding (Fig. \ref{fig:framework} (c)), making points with different geometry more distinguishable. In the encoder, geometric patterns can be learnt from different scales, thus complex geometric pattern can be represented by the combination of geometrical features of different scales.

\section{Experiments}

The experiments can be divided into two parts. We demonstrate the performance of our method and compare it with other state-of-the-art methods on ScanNet v2~\cite{dai2017scannet} and S3DIS Area-5~\cite{armeni20163d} for scene semantic segmentation task, respectively. Then, intensive ablation studies are conducted. We take the mean intersection-over-union (mIoU) over categories as our metric like many previous works~\cite{wu2019pointconv}. 

\subsection{Scene Semantic Segmentation}
\label{subsec: semantic}

\subsubsection{Dataset.} In scene semantic segmentation task, we evaluate our method on ScanNet v2~\cite{dai2017scannet} and S3DIS~\cite{armeni20163d}. In ScanNet v2, there are totally $1,201$ scanned scenes for training and $312$ scenes for validation. Additionally, another 100 scenes are provided as the testing samples, and there are 20 different categories. Following~\cite{wu2019pointconv}, we randomly sample $3m\times 1.5m \times 1.5m$ cubes from rooms with 8,192 points as the training samples, and test over the entire scan. 
In S3DIS, there are six indoor areas including 271 rooms from three different buildings. Each point is annotated with a corresponding label from 13 categories. We split points by room and sample all rooms into $0.5m \times 0.5m$ blocks with $0.25m$ padding. Like experiment setting used in previous works~\cite{qi2017pointnet,li2018pointcnn}, we split Area 5 as the test set and use others for training. In the training areas, 4,096 points are sampled for each block and all points in the testing areas are used for testing block-wisely.

\subsubsection{Implementation.} In our method, we take an efficient way to implement the weight computation and feature aggregation using matrix multiplication like PointConv~\cite{wu2019pointconv}. Therefore, we take PointConv as our baseline, but we do not use density information during feature extraction because it have limited improvement in performance. 

In the BPM, to automatically annotate the target boundary points for each input point cloud, points with more than $40\%$ of $32$ neighbors not belonging to the same category are assumed to be boundary points. Then, because boundary points are predicted based on neighborhood information, and color information is highly related to boundary prediction, we take the variance of color features of $32$ neighbors as the aggregated feature for each point and further predict the boundary points. After predicting boundary points, we build an encoder-decoder network based on the boundary-aware GEM and take both the color and coordinate information as its input. Our model is trained by Adam optimizer with batch size 8 for ScanNet and batch size 12 for S3DIS on a GTX 1080Ti GPU. Also, we analyze the number of the ground truth of boundary and non-boundary points in different scenes.  Accordingly, for ScanNet, $w_1$ and $w_2$ used in $\mathcal{L}_{BPM}$ are 1 and 10, and for S3DIS, $w_1$ and $w_2$ are 1 and 2.

\subsubsection{Results.}
\begin{table}[t]
\centering
\begin{tabular}{lc}  
\toprule
Method  &  mIoU  \\
\midrule
PointNet++ ~\cite{qi2017pointnet++} & 33.9 \\
PointCNN ~\cite{li2018pointcnn} & 45.8 \\
3DMV ~\cite{dai20183dmv} & 48.4 \\ 
PointConv ~\cite{wu2019pointconv} & 55.6 \\
TextureNet ~\cite{huang2019texturenet} & 56.6 \\
HPEIN ~\cite{jiang2019hierarchical} & 61.8 \\
SegGCN~\cite{lei2020seggcn} & 58.9\\
SPH3D-GCN~\cite{lei2020spherical} & 61.0 \\
FusionAwareConv~\cite{zhang2020fusion} & 63.0	\\
\midrule
Ours &\textbf{63.5} \\
\bottomrule
\end{tabular}
\caption{Semantic segmentation results on ScanNet v2.}
\label{tab:semantic}
\end{table}

\begin{table}
\begin{center}
\begin{tabular}{lc}
\toprule
Method & mIoU \\ 
\midrule
PointNet~\cite{qi2017pointnet} & 41.09 \\
PointCNN~\cite{li2018pointcnn} & 57.26 \\
SPGraph~\cite{landrieu2018large} & 58.04 \\ 
PCCN~\cite{wang2018deep} & 58.27 \\ 
ASIS~\cite{wang2019associatively} & 53.40 \\ 
ELGS~\cite{wang2019exploiting} & 60.06 \\ 
PAT~\cite{yang2019modeling} & 60.07 \\ 
SPH3D-GCN~\cite{lei2020spherical} & 59.5 \\
GridGCN~\cite{xu2020grid} & 57.75 \\ 
JSNet~\cite{zhao2020jsnet} & 54.50 \\ 

\midrule
Ours & \bfseries 61.43 \\ 
\bottomrule
\end{tabular}
\end{center}
\caption{Semantic segmentation results on S3DIS evaluated on Area 5 (Fold \#1).}
\label{tab:s3dis}
\end{table}

\begin{figure}[t]
    \centering
    \includegraphics[width=\linewidth]{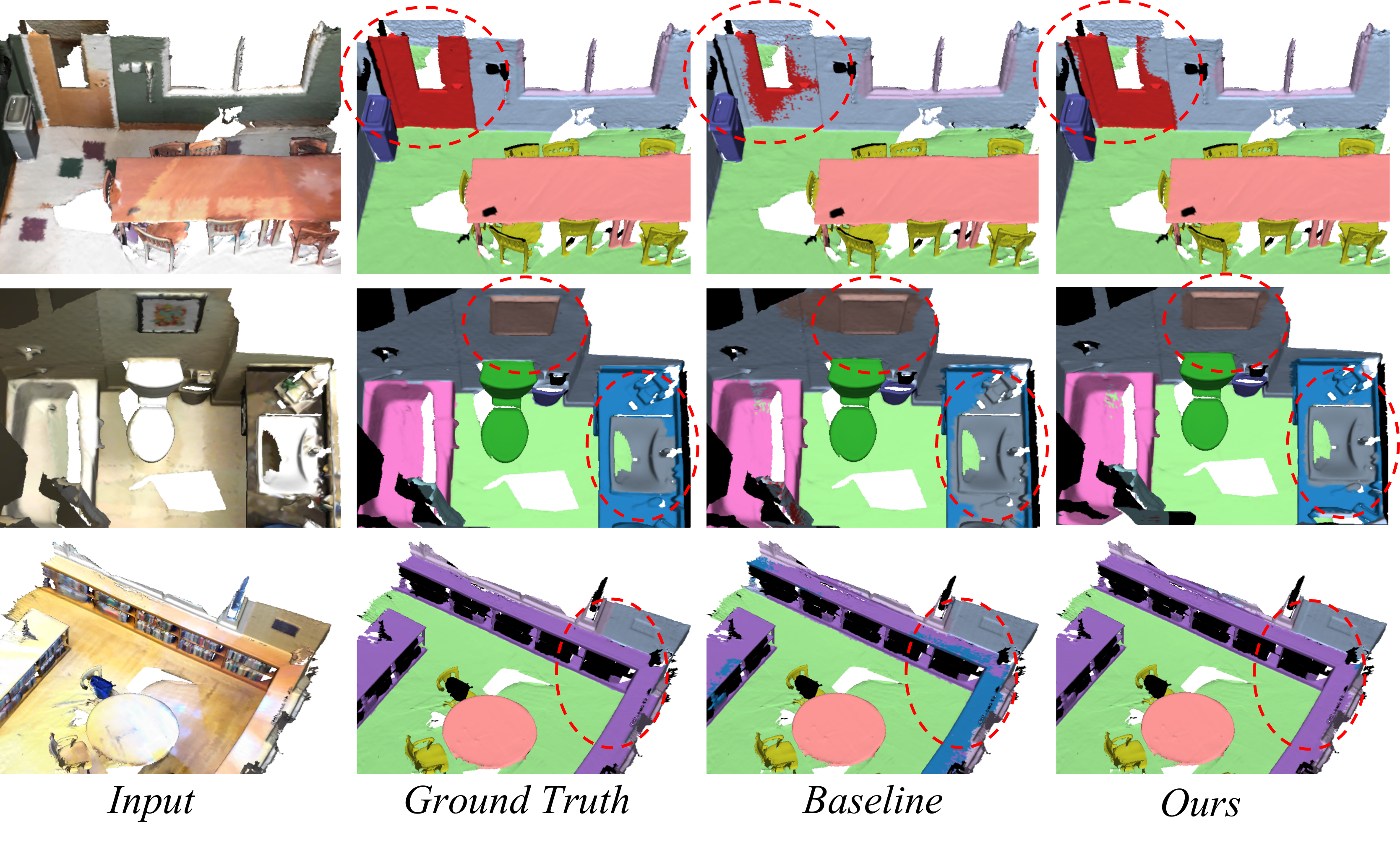}
    \caption{Visualization results of scene semantic segmentation on ScanNet v2. The images from left to right are separately the input scene, ground truth of segmentation, results predicted by PointConv and our method.}
    \label{fig:scannet_vis}
\end{figure}

For ScanNet v2, we report the mean IoU (mIoU) over categories in Table \ref{tab:semantic}, where we have achieved mIoU of $63.5\%$. It shows our method has outperformed lots of state-of-the-art competitors. Fig. \ref{fig:scannet_vis} visualizes scene semantic segmentation result of PointConv and our method. Misclassification is easy to appear in the transition area of two adjacent objects. For example, in the second row third column, points of ``wall'' category are predicted as the ``picture'' that is adjacent to the wall, leading to the poor contour of the picture. By contrast, benefiting from the boundary awareness, our network perform well in this transition area. 

\begin{figure}[t]
    \centering
    \includegraphics[width=\linewidth]{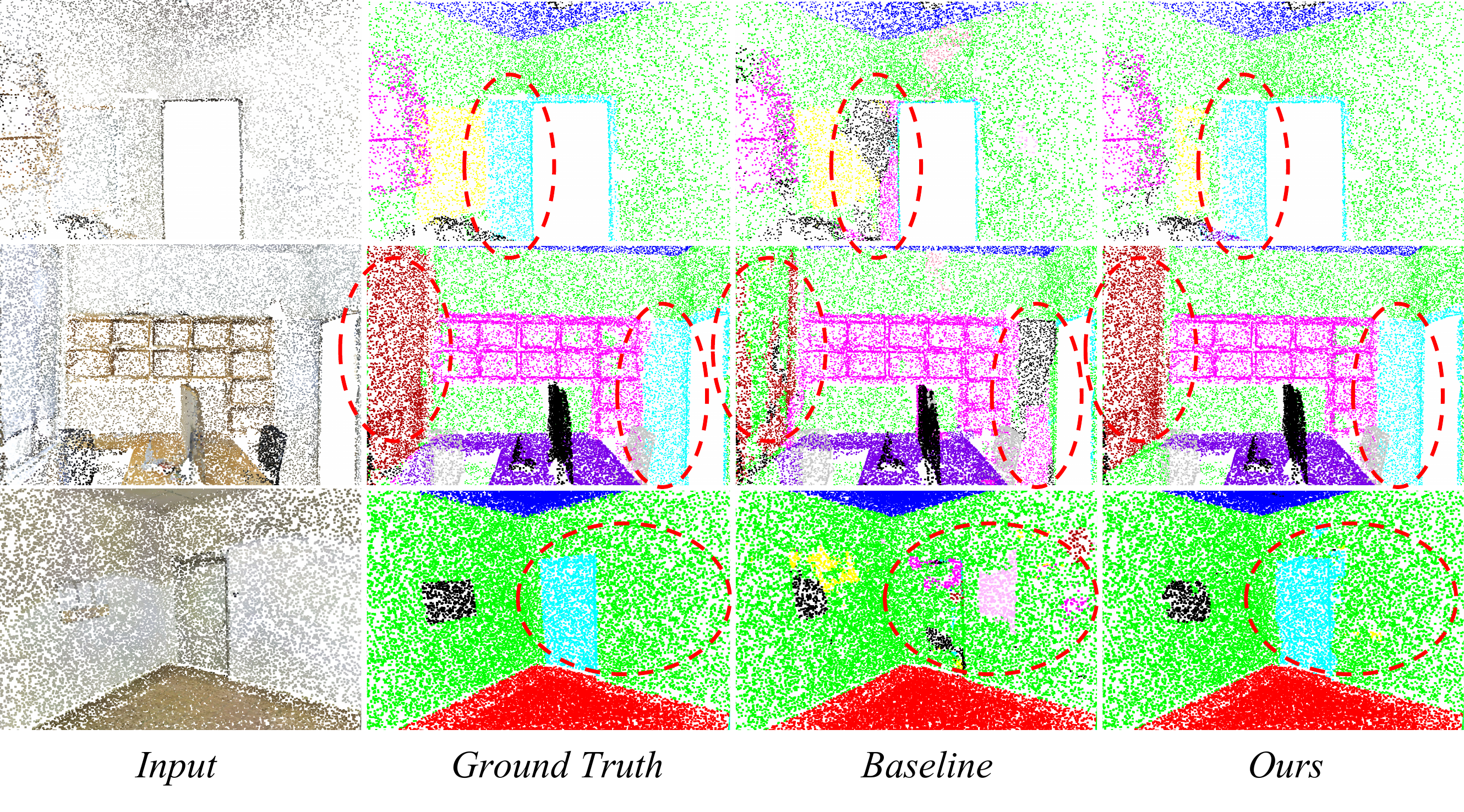}
    \caption{Visualization results of scene semantic segmentation on S3DIS (Area \#5). The images from left to right are separately the input scene, ground truth of segmentation, results predicted by PointConv and our method.}
    \label{fig:s3dis_vis}
\end{figure}

For S3DIS, we report the mIoU over categories in Table \ref{tab:s3dis}. We achieve $61.43\%$ in mIoU on this benchmark which has better performance than many state-of-the-art competitors. Also, we visualize our segmentation results in Fig. \ref{fig:scannet_vis}. As can be seen in this figure, the segmentation results obtained by our method have better contour thanks to the Boundary-aware GEM for local feature extraction.

\subsection{Ablation Study}
\label{subsec:ablation}

In this section, we conduct more ablation studies to support our contributions. Because we can only submit one final result to the testing benchmark server of ScanNet, more ablation studies are conducted on the validation set of ScanNet.

\begin{table}[t]
    \centering
    \begin{tabular}{lc}
    \toprule
    Method & mIoU\\
    \midrule
    Baseline &  58.9\\
    \midrule
    Baseline w/ GCO & 60.4\\
    BAGEM w/o GCO & 60.9\\ 
    \midrule
    Boundary Augmented & 61.8\\
    \midrule
    Proposed method &  \textbf{63.4}\\
    \bottomrule
    \end{tabular}
    \caption{Results on the ablation study of boundary-aware GEM. BAGEM represents boundary-aware GEM, Boundary Augmented means to use a strategy to augment the contribution of boundary point in feature aggregation.}
    \label{tab:ablation}
\end{table}
    
\begin{table}
    \centering
    \begin{tabular}{lcc}
    \toprule
    Geo. Encoding &  mIoU & FLOPs~\\
    \midrule
    KC &  60.8 & 26.93G~\\
    \midrule
    KC (H) &  61.0 & 3.87G~\\
    \midrule
    GCO (2) &  61.1 & 3.65G~\\
    GCO (6) &  61.7 & 4.20G~\\
    \midrule
    {Proposed method} &  \textbf{63.4} & 3.73G~\\
    \bottomrule
    \end{tabular}
    \caption{Results on the ablation study of geometry encoding. KC means replacing GCO with Kernel Correlation~\cite{shen2018mining}. KC (H) means use a light-weight version of KC hierarchically. The number after GCO means the number of vectors in the kernel.}
    \label{tab:kernel_abla}
\end{table}

\subsubsection{Effectiveness of boundary-aware GEM and GCO.}

To show the effectiveness of boundary-aware GEM and the GCO, we conduct more ablative experiments. First, we only use MLPs to simulate the 3D convolution kernel like PointConv~\cite{wu2019pointconv} and treat it as our baseline. Next, we simply introduce GCO into the baseline to validate the effectiveness of GCO. Then, we use boundary-aware GEM without GCO to build the network and prove the effectiveness of boundary-aware strategy. Finally, we report the result of our method on validation dataset. The results are shown in Table \ref{tab:ablation}, in which we can see both of them can improve the performance on semantic segmentation task.

\subsubsection{Strategy on Boundary Utilization.}

In our method, we attempt to prevent the propagation of features for points on the boundary for local feature extraction. Meanwhile, BSANet~\cite{zhao2019multi} proposed to emphasize the features near boundary. So we try to enhance the influence of points on the boundary during local feature aggregation by:
\begin{equation}\label{modified_local_aggregation}
    f_{en\_\,l}=\sigma(\mathcal{A}(\{\phi(r_{ij})\cdot\mathcal{M}((2-\hat{g_j})\cdot f_{p_j})\})),
\end{equation}
where $x_j\in\mathcal{N}(p_i)$ and other settings are the same as our proposed method. Recall the Eq. (\ref{local_aggregation}), $\hat{g_j}$ is $0$ if $p_j$ is on the boundary. Given the Eq. (\ref{modified_local_aggregation}), more emphasis is imposed on boundary points. The result is shown in Table \ref{tab:ablation} (``Boundary Augmented''), achieving $61.8\%$ in mIoU, which is better than not using boundary information. However, compared with our proposed method, it decreases the mIoU by $1.6\%$, that means preventing the propagation of features on the boundary is more effective than emphasizing the features near boundary.

\subsubsection{Performance of GCO compared with KCNet.}

Both KCNet and GCO utilize a vector-set kernel to represent localgeometric pattern. Compared with KCNet~\cite{shen2018mining}, our geometric features are more sensitive to direction rather than position, thus less sensitive to density of points. Additionally, we take a strategy to extract geometric features hierarchically with light-weight geometric convolution to learn complex pattern rather than a heavy-weight module to extract geometric features within one layer. 

To show our advantages and give fair comparisons, we first replace GCO with Kernel Correlation (KC) proposed in KCNet with the settings same as KCNet (corresponding to the first row in Table \ref{tab:kernel_abla}). More specifically, the KC is only employed in the first layer with $16$ learnable kernel vectors. Moreover, following the settings of our proposed method, we use a light-weight version of KC, that reduces the learnable kernel vectors from $16$ to $3$, to extract geometric information hierarchically (denoted by KC (H) in Table \ref{tab:kernel_abla}). Also, the extra computational cost is illustrated for these geometry encoding methods in terms of FLOPs. Comparing the row $2$ vs. row $1$ in Table \ref{tab:kernel_abla}, it is shown that using light-weight version of KC to extract geometric features hierarchically obtains better performance than using heavy-weight KC in one layer like KCNet. In addition, using light-weight version of KC decreases the computation drastically. More importantly, keeping other settings the same and using GCO can further increase the mIoU by $2.4\%$. Compared with the light-weight version of KC, GCO has a much simpler form and require less computation resource.

\subsubsection{Number of vectors in geometric kernel.}

In our implementation, we utilize a kernel unit with a set of only three $3$-D vectors.
To check whether a kernel with more or less vectors can learn geometric pattern better, we separately take a kernel with six and two $3$-D vectors and other settings are the same. The results are shown in Table \ref{tab:kernel_abla}, where taking six $3$-D vectors as kernel achieves $61.7\%$ in mIoU which may be caused by overfitting. It also proves our claim that a kernel with three vectors are enough to learn 3D geometry in a hierarchical way and a kernel with less than three vectors is not able to learn 3D geometry, thus has worse performance.

\begin{table}
    \centering
    \begin{tabular}{lc}
    \toprule
    Method &  mIoU \\
    \midrule
    No boundary information & 60.4 \\
    \midrule
    Random flip & 62.4 \\
    Exchange neighboring pair &  61.8~\\
    \midrule
    No perturbation & \textbf{63.4}\\
    \bottomrule
    \end{tabular}
    \caption{The result of perturbing the predicted boundary point, which shows the robustness of our method.}
    \label{tab:robustness}
\end{table}

\subsubsection{Robustness to boundary prediction error.}
Furthermore, we also conduct ablation study to show the robustness to prediction error introduced by BPM. First, we randomly flip 3\% points on prediction results.  
We report the results on Table \ref{tab:robustness} and 62.4\% mIoU is achieved on the validation set of ScanNet. We think that 3\% is enough because the number of randomly flipped points is comparable to the number of  target boundary points. Second, we select 5\% of the predicted boundary points and exchange the label of each point with its nearest neighbor, making the boundary points shifted by one point and 61.8\% mIoU was achieved. Both outperform the network without predicted boundary, showing the robustness to errors in boundary prediction.

\section{Conclusion}

In this paper, we propose a boundary-geometry aware segmentation method including a Boundary Prediction Module (BPM) and boundary-aware Geometric Encoding Module (boundary-aware GEM) with Geometric Convolution Operation (GCO). The BPM supervised by the dynamically generated target boundary can predict the boundary points in the point cloud. In the boundary-aware GEM, the predicted boundary will guide the feature aggregation by ignoring the contribution of boundary points when collecting features of neighboring points. To exploit the geometry information, we propose the GCO to recognize geometry patterns at different scales and provide extra geometry information. Our proposed method achieves state-of-the-art performance on both ScanNet and S3DIS dataset for 3D semantic segmentation.

\section{ Acknowledgments}
We thank for the support from National Natural Science Foundation of China (61972157, 61772524, 61876161, 61902129), Zhejiang Lab (No. 2020NB0AB01), Natural Science Foundation of Shanghai (20ZR1417700), National Key Research and Development Program of China (2019YFC1521104, 2020AAA0108301), Shanghai Municipal Commission of Economy and Information  (XX-RGZN-01-19-6348). Jingyu Gong is also supported by Wu Wen Jun Honorary Doctoral Scholarship, AI Institute, Shanghai Jiao Tong University.

\bibliography{boundary}
\end{document}